%% file: main_neurips.tex
\documentclass{article}

\usepackage[preprint]{neurips_2025}

\usepackage[utf8]{inputenc} 
\usepackage[T1]{fontenc}    
\usepackage[colorlinks=true]{hyperref}       
\usepackage{xcolor}         
\definecolor{pAlgae}{RGB}{87,155,225}
\hypersetup{
     colorlinks=true,
     linkcolor=blue,
     filecolor=blue,
     citecolor=pAlgae,      
     urlcolor=cyan,
     }
     
\usepackage{url}            
\usepackage{booktabs}       
\usepackage{amsfonts}       
\usepackage{nicefrac}       
\usepackage{microtype}      
\usepackage{graphicx}
\usepackage{makecell}
\usepackage{tabularx}
\usepackage{amsmath}
\usepackage{bm}

\usepackage[frozencache,cachedir=.]{minted}
\usepackage{xcolor}
\usepackage[most]{tcolorbox}
\tcbuselibrary{listings, breakable, skins}
\definecolor{promptpurple}{RGB}{225, 230, 255}
\newtcolorbox[list inside=mybox,auto counter]{fancypromptbox}[2][]{
  enhanced,
  float=htb,
  colback=promptpurple,
  colframe=black,
  boxrule=0.5pt,
  arc=4pt,
  outer arc=4pt,
  boxsep=5pt,
  left=5pt,
  right=5pt,
  top=5pt,
  bottom=5pt,
  overlay={
    \node[anchor=north west, fill=black, text=white, font=\bfseries, rounded corners=2pt, inner sep=2pt, font=\sffamily]
    at ([xshift=15pt,yshift=7pt]frame.north west) {#2};
  },
  #1
}

\newcommand{\pyprefix}{>{}>{}>}

\title{Replacing thinking with tool usage enables reasoning in small language models}

\author{%
  Corrado Rainone\thanks{Equal contribution}\\
  Qualcomm AI Research\thanks{Qualcomm AI Research is an initiative of Qualcomm Technologies, Inc.} \\
  \texttt{crainone@qti.qualcomm.com} \\
  \And
  Tim Bakker\footnotemark[1] \\
  Qualcomm AI Research\footnotemark[2] \\
  \AND
  Roland Memisevic \\
  Qualcomm AI Research\footnotemark[2] \\
}

\begin{document}

\maketitle

\begin{abstract}
  Recent advances have established a new machine learning paradigm based on scaling up compute at inference time as well as at training time. In that line of work, a combination of Supervised Fine-Tuning (SFT) on synthetic demonstrations and Reinforcement Learning with Verifiable Rewards (RLVR) is used for training Large Language Models to expend extra compute during inference in the form of ``thoughts'' expressed in natural language. 
  In this paper, we propose to instead format these tokens as a multi-turn interaction trace with a stateful \emph{tool}. At each turn, the new state of the tool is appended to the context of the model, whose job is to generate the tokens necessary to control the tool via a custom DSL. We benchmark this approach on the problem of repairing malfunctioning Python code, and show that this constrained setup allows for faster sampling of experience and a denser reward signal, allowing even models of size up to 3B parameters to learn how to proficiently expend additional compute on the task.
\end{abstract}

\section{Introduction}
It was first observed in~\citet{Wei2022-cotfirst} that prompting capable Language Models (LMs) to ``think'' step-by-step, rather than to provide an immediate answer, could lead to significant performance boosts. Currently, Large Reasoning Models (LRM)~\citep{Fengli2025-reasoningsurvey}, which generate additional tokens in the form of Chains-of-Thought (CoTs) in order to improve their performance on a question or task, have effectively become a new AI paradigm \citep{openai-o1, DeepSeek-AI2025-deepseek}. Since LRMs aim to improve their performance on a task by expending additional compute at inference time, they are part of a class of approaches referred to as Test-Time Scaling (TTS) or Inference-time Compute (ItC) methods; this class also includes approaches that steer the decoding process via some form of (e.g., tree) search, with outputs scored by either an engineered reward function or a verifier model~\citep{hf-scalingtesttimecompute, Snell2024-testscalebthantrainscale, openai-o1, Liu2025-intext}, an idea that can ultimately be traced all the way back to the `Alpha-' series of works~\citep{Silver2016-alphago, Silver2017-alphazero}. TTS methods, including LRMs, are often taken to embody ``System-2'' thinking~\citep{Ji2025-itcsystem12}, the slow and deliberate thinking processes used by humans to plan and adapt to previously unseen situations.

LRM training is typically done at least in part on \emph{experience}; i.e., based on their own outputs rather than on external demonstrations~\citep{Zelikman2022-star, Wang2024-reasoningdpo, Setlur2025-rlbetterforitc, DeepSeek-AI2025-deepseek, Hu2025-openreasoner, Chu2025-stfvsrl}. Recently, this has taken the form of on-policy Reinforcement Learning (RL) on tasks that enable the handcrafting (as opposed to neural modeling) of a reward function. Such \emph{verifiable} tasks are mostly sourced from the domains of math and coding. This paradigm of Reinforcement Learning with Verifiable Rewards (RLVR) was compellingly demonstrated by~\citet{DeepSeek-AI2025-deepseek}, who trained a model to reason via on-policy RL on its own CoTs. Regardless of whether reasoning is learned from experience or from expert demonstrations~\citep{Ye2025-limo, Muennighoff2025-ks} the additional tokens generated at test time typically take the form of reasoning steps described in natural language, which~\citet{Feng2025-retool} refer to as ``text-based reasoning''. Although this parameterization has the benefit of generality due to the representative power of natural language, it also comes with pitfalls. Only large models have so far been ``bootstrapped'' on their own CoTs, as was done in~\cite{DeepSeek-AI2025-deepseek}. For smaller models, one may attempt to distill the reasoning capabilities of larger models and subsequently attempt on-policy RL~\citep{Quy-Anh2025-whatworks}; one may moreover attempt to unlock reasoning behavior in small models via simple Supervised Fine-Tuning (SFT) on expert demonstrations~\citep{Ye2025-limo, Muennighoff2025-ks}. However, such methods are yet unproven to work for models of fewer than 10B parameters, or suffer from instabilities that preclude stable and continuous learning from experience~\citep{Quy-Anh2025-whatworks}. The action space in which any LM agent operates is enormously large, and the reward signal of RLVR, usually based on task outcome~\citep{DeepSeek-AI2025-deepseek, Hu2025-openreasoner, Quy-Anh2025-whatworks}, is only available once the model has completed its reasoning and produced an answer. Unless these challenges are mitigated, successful RLVR inevitably requires a strong exploration prior, which small LMs struggle to provide.

In this paper, we propose a different parameterization of ``thinking'' tokens, which reduces the size of the action space and allows for a denser reward signal. We benchmark our proposal on the verifiable task of repairing malfunctioning Python code, and demonstrate that it enables extending the RLVR paradigm to models as small as 1B in size using only Low-Rank adaptation~\citep{Hu2021-lora}. In our proposal, we format ``thinking'' tokens as the tracing of a series of interactions between the reasoner and an external tool, which we refer to as a Chain-of-Edits (CoE). 
We make the following contributions:
\begin{itemize}
\item We propose to parameterize ``thinking'' tokens as an interaction trace between a model and a text editor, wherein Python code can be edited via a custom DSL and executed against a set of unit tests to obtain feedback. We refer to these traces as ``Chains-of-Edits'' (CoE).
\item In order to teach small language models to use such CoEs, we propose a pipeline that consists of SFT and RLVR stages. These training stages use only Low-Rank adaptation~\citep{Hu2021-lora}.
\item We apply this pipeline on a trio of SLMs ranging in size from 1B to 8B and compare it with a baseline method that aims to induce reasoning behavior in the form of text-based CoTs. We show that for sizes up to 3B, the CoE approach most successfully elicits reasoning behavior, i.e., the model performs better when using CoE than when few-shot prompted for a direct answer, while the text-based CoT approach fails to induce such an improvement.
\end{itemize}

\section{Related work}

\paragraph{Training small-size reasoners}
Building small reasoning models (i.e., with fewer than 10B parameters) may be achieved by distillation from a larger and more capable teacher model~\citep{Mukherjee2023-orca1, Mitra2023-orca2, DeepSeek-AI2025-deepseek}, or by coupling a policy model with a verifier model that steers the generation process~\citep{hf-scalingtesttimecompute, Xinyu2025-rstar}. Training a small reasoner from scratch may also be feasible but presents, especially when RL is involved, challenges due to unstable training and limited context length~\citep{Quy-Anh2025-whatworks}. In~\citet{DeepSeek-AI2025-deepseek}, it is explicitly stated that training a large teacher model with RL and then distilling its reasoning into a small student is preferable to directly training the student. In this work, we propose and test a recipe for directly training a small reasoning model. Our recipe involves a combination of SFT on synthetic (but not teacher-generated) demonstrations, RLVR, and the affordance of an external tool and verifier feedback. We observe that this recipe mitigates some of the difficulties observed with the exclusively ``free-form reasoning'' approach of \citet{Quy-Anh2025-whatworks}. Although we integrate verifier feedback into the context during inference , our decoding procedure remains entirely auto-regressive, in contrast to~\citet{hf-scalingtesttimecompute, Xinyu2025-rstar}.

\paragraph{Code generation and repair}
Our aim of unlocking ItC scaling for coding is shared by~\citet{Ulyana2024-wy}, who propose to train LMs to sequentially generate code as a sequence of error-free diffs, rather than in a single pass; in our work, we focus on code repair and editing rather than code generation, and we equip the LMs with an editing environment and execution feedback. Teaching LMs to self-refine -- with or without code execution feedback -- has been explored in several other works, such as~\citet{Gupta2020-sed, Yang2023-intercode, Chen2023-selfdebug, Madaan2023-selfrefine, Gehring2024-rlef}. \citet{Gupta2020-sed} design a code-generation pipeline in which a LM generates an initial code draft that it iteratively revises by outputting one edit operation per token, based on feedback in the form of an execution trace; as in the present work, the LM is first trained on synthetic corruptions of code snippets and is confined to small-scale revisions. \citet{Gehring2024-rlef} use RL to teach capable LMs to refine their code generations over multiple turns while benefiting from execution feedback; however, they limit their framework to three turns and the model is required to output fully revised code every turn. Solving coding problems at scale in an ``agentic'' manner is the focus of other works as well, such as~\citet{Jimenez2023-swebench, Yang2023-intercode, Guo2024-edbench, Yang2024-sweagent,  Wei2025-swerl}. Although we occasionally refer to our framework as ``agentic'', the task we focus on is not \emph{ab-initio} agentic in nature, and our main focus remains on enabling ItC behavior \emph{via} an agentic workflow.

\paragraph{Tool-usage for Language Models} 
Teaching LMs how to properly use tools either via SFT or in-context learning is the focus of many works, among them \citet{Schick2023-toolformer, Hao2023-toolken, Erdogan2024-tinyag, Li2025-start}. A common theme, which is also present in our work, is to aid LMs in avoiding common pitfalls, such as hallucinations, by having them not rely exclusively on their own context and internal representations. In other works, such as~\citet{Erdogan2024-tinyag}, the task is simply to learn proper use of a suite of tools. Taking advantage of tools as a way to build ItC structures is the subject of ref.~\citet{Yang2022-procclon}, which proposes tool-usage traces as a means to build expert CoTs via procedure cloning, although in that work the LM is not allowed to use those tools during inference. A recent reference -- that is quite close in spirit to ours -- is~\citet{Feng2025-retool}, which employs a combination of SFT and RL to train a tool-based reasoner and benchmark it against purely ``text-based'' reasoners.

\section{The Chain-of-Edits}
\begin{fancypromptbox}[label=box:state_example1]{Box \thetcbcounter: Example of a state, complete with line markers and execution feedback.}{\obeylines
\begin{minted}{python}
L  1 
L  2 def max_sub_array_sum(arr,n,k): 
L  3     sum = 0 
L  4     max_sum = 0
L  5     for i in range(n): 
L  6         sum = sum + arr[i]
L  7         if (i >= k): 
L  8             sum = sum - arr[i-k] 
L  9         max_sum = max(max_sum, sum) 
L 10     return max_sum
\end{minted}
***
You defined the function with the wrong number, or wrong type, of arguments. Here is the stack\_trace:
Traceback (most recent call last):
\begin{verbatim}
  File "/app/run/code_edits/verifiers.py", line 137, in __call__
    exec(test, global_names, global_names)
  File "<string>", line 1, in <module>
\end{verbatim}
\begin{minted}{python}
TypeError: max_sub_array_sum() missing 1 required positional argument:
 'k';
\end{minted}
}
\end{fancypromptbox}
We now outline the characteristics of our CoE approach. Like CoTs, CoEs are meant to be extra tokens that a model generates to solve a problem step-by-step. In contrast to CoTs, we define these tokens as the tracing of a series of interactions between an LM agent and a stateful environment; the agent alternates between observing the state of the environment and issuing an action, which is then executed by the environment to locally edit the state, over multiple turns of interaction. In this process, the agent does not observe or produce tokens interpretable as free-form reasoning. This way the agent only has to generate a few tokens during inference, speeding up the sampling of experience during RL and concentrating exploration. It also does not have to keep track of the environment's state, yet the full task still decomposes into a series of \emph{local} steps, each of which can be scored by a \emph{dense} reward function. 
We believe that these improvements w.r.t the CoT approach will enable reasoning-like capabilities in weaker and smaller models.

\subsection{The scratchpad and its DSL\label{subsec:scr}}
The environment that the agent is provided with consists of a text editor with code execution capabilities; we refer to it as both an environment and a tool. Like an environment, it keeps a \emph{state} that consists of the code snippet that resides in the editor, as well as the execution feedback that results from an execution attempt of this snippet. Like a tool, it is meant to help the agent solve a code repair task, which the agent does by taking \emph{actions} that modify the editor's state.

\paragraph{Verification and execution feedback}
We define a \emph{code repair} task as consisting of the following elements:
\begin{itemize}
\item A natural-language description of a coding task.
\item A set of unit tests against which a candidate solution to the task should be checked. The task is considered solved when all unit tests pass.
\item A piece of Python code that fails one or more unit tests. This could be due to wrong syntax, inconsistent variable names, or more subtle issues such as not satisfying the task's precise requirements.
\item A ground-truth code snippet that solves the task.
\end{itemize}
Therefore, a code repair task is by definition \emph{verifiable}. For each task, a text editor with verification capabilities is instantiated. The \emph{state} of the editor is given by its code content, followed by the execution feedback that results from running that content against the task's unit tests. The content is marked with line numbers, which we observed to be helpful for the agent. A separator (a sequence of three asterisks, \texttt{\small ***}) separates the code content from the associated feedback. In Box~\ref{box:state_example1}, we provide an example of the state of the text editor (other examples may be found in the Appendix). We stipulate that when all unit tests pass and the task is solved, the execution feedback is empty.

\paragraph{The editing DSL\label{par:dsl}}
The LM agent can modify the content of the scratchpad by issuing \emph{editing commands} in a domain-specific language (DSL) of our design. Each time this happens, the edited content is executed against the unit tests. This results in execution feedback that, together with the updated scratchpad contents, forms the updated state.
We designed the DSL to encourage small-scale, atomic edits of Python code. Each command consists of an instruction formatted in uppercase, which can be followed by one or more arguments. We refer to Table~\ref{tab:DSL} for an outline and example of each command.

\begin{table}
    \caption{Outline of the DSL commands that the model may use for editing the content of the scratchpad. The model is restricted to using a single command every interaction turn.}
    \centering
    \begin{tabular}{ccc}
        \toprule\toprule
        \textbf{Command} & \textbf{Description} & \textbf{Example} \\
        \toprule
        \texttt{\small ADDL \#L \pyprefix<python>} & \makecell{\tiny At the line number indicated by the first argument,\\ \tiny add a new line consisting of <python>} & \texttt{\small ADDL 4 \pyprefix    return None} \\
        \midrule
        \texttt{\small REPL \#L \pyprefix<python>} & \makecell{\tiny Replace line at the line-number indicated by first argument,\\ \tiny with a line consisting of <python>} & \texttt{\small REPL 4 \pyprefix    return None} \\
        \midrule
        \makecell{\texttt{\small REPW \pyprefix<python>}\\ \texttt{\small \pyprefix<python> \#L}} & \makecell{\tiny Replace <python> at second argument\\\tiny with <python> at the third argument,\\\tiny at line in the first argument} & \texttt{\small REPW \pyprefix  ronge \pyprefix range 3} \\
        \midrule
        \texttt{\small DELL \#L} & \makecell{\tiny Delete line at the line-number indicated by the argument.} & \texttt{\small DELL 3} \\
        \midrule
        \texttt{\small EXIT} & \makecell{\tiny Terminate trace/episode} & \\
        \toprule
    \end{tabular}
    \label{tab:DSL}
\end{table}

\section{Training pipeline}
Having defined the components of a CoE, we now outline how we fine-tune a language model to use CoEs to solve code repair tasks over multiple turns of interaction. Our pipeline consists of two stages;
\begin{enumerate}
\item Supervised Fine-Tuning (SFT) on a dataset of synthetic demonstrations of CoE usage; 
\item Reinforcement Learning with Verifiable Rewards (RLVR) on a suite of code repair tasks.
\end{enumerate}
In both of these stages, we only make use of LoRA~\citep{Hu2021-lora} for gradient updates. In Section~\ref{subsec:mbppr} we outline how our code repair tasks are constructed.


\subsection{Supervised Fine-Tuning on CoE demonstrations}
This first step serves to teach the model how to fix a broken piece of code by interacting with the text editor defined in Section~\ref{subsec:scr}, by exposing it to the syntax of the editor's DSL. Here, the model may also learn how edits relate to the previous state of the code and verifier output. We achieve this by generating a dataset of synthetic CoE traces that demonstrate full state-action trajectories (in tokens) of code-repair tasks.

\subsubsection{Generating CoE demonstrations} \label{sec:synth}
Generating CoE demonstrations is not trivial. One option is to train human programmers on DSL usage and ask them to perform the task of repairing corrupted code snippets by hand. Such a procedure would require expensive and time-consuming work by human experts. Instead, we opt to develop an automated pipeline for generating synthetic demonstrations. The pipeline consists of three steps: 

\begin{enumerate}
    \item Starting from an initial and correct code snippet for a given task, randomly generate a corruption that can be inverted by any of the provided DSL commands (e.g., deleting a line) and apply this corruption to the code snippet.
    \item Apply step 1 any desired number of times $c$, until the original code snippet is sufficiently corrupted (e.g., after $c=C$ steps, for some $C$).
    \item Create a CoE demonstration by \emph{reversing} this process: starting from the corrupted code snippet at step $c=C$, sequentially reverse the applied corruptions by using the DSL commands. Since the corruptions are invertible by construction, this recovers the original (correct) ground-truth code snippet from step $c=0$ (before the first corruption). 
\end{enumerate}

\begin{fancypromptbox}[label=box:mbpps_train]{Box \thetcbcounter: Example of a CoE demonstration prefix prompt for the \texttt{count\_Divisors} task.}{\obeylines
You are an expert Python programmer whose goal is to fix all mistakes in a code snippet. You may interact with the code snippet only by applying the provided DSL commands. Valid DSL command templates are:
\verb|`### DELL <line_number>`| to delete the line at the specified line number.
\verb|`### ADDL <line_number> >>><line_content>`| to add a line at the specified line number with the specified content.
\verb|`### REPL <line_number> >>><line_content>`| to replace the line at the specified line number with the specified content.
\verb|`### REPW <line_number> >>><string_to_replace> >>><string_to_insert>`| to replace all specified strings in the line at the line number with the new string.
Here is your task: Write a python function to check whether the count of divisors is even or odd.
Your code should pass these tests:
\begin{minted}{python}
assert count_Divisors(10) == "Even"
assert count_Divisors(100) == "Odd"
assert count_Divisors(125) == "Even"
\end{minted}
Below is an initial malfunctioning code snippet to fix:
\begin{minted}{python}
L  1 import math
L  2 def count_Divisors(n) :
L  3     count = 0
L  4     for i in range(1, (int)(math.sqrt(n)) + 2) :
L  5         if (n % i == 0) :
L  6             if( n // i == i) :
L  7                 count = c=unt + 1
L  8             else :
L  9                 count = count + 2
L 10     if (count % 2 == 0) :
L 11 def remove_nrgs(num_list):
L 12     else :
L 13         return ("Odd")
L 14 def profit_amount(actual_cost,sale_amount):
\end{minted}
***
The syntax of the proposed solution was not correct. Here is the stack trace:
  File "<string>", line 11
\mint{python}{    def remove_nrgs(num_list):}
IndentationError: expected an indented block after 'if' statement on line 10;}
\end{fancypromptbox}

The CoE demonstration trace is constructed as a sequence of states and actions. A prefix prompt specifies the task that the repaired code snippet should perform and provides a number of unit tests to pass. The prompt also explains the DSL commands; see Box~\ref{box:mbpps_train} for an example prompt. The prefix prompt is followed by the first state, which contains the fully corrupted code snippet $s_C$ (equipped with line markers) and the corresponding execution feedback; see Box~\ref{box:state_example1} for an example. The rest of the demonstration consists of a sequence of interleaved DSL commands (actions) and the resulting scratchpad content with execution feedback (next states) that recover the original snippet $s_0$. The final action is a special \texttt{\small EXIT} action that signals the end of the demonstration. 
\begin{table}
  \caption{The corruption types used in automated generation of \emph{Chain-of-Edits} demonstrations. Every corruption type is matched with a DSL command that can reverse the corruption.}
  \label{tab:corruptions}
  \centering
  \begin{tabular}{ll}
    \toprule
    Corruption type & Reversing DSL command \\
    \midrule
     Delete a line & \texttt{ADDL} (add line) \\
     Add a line & \texttt{DELL} (delete line) \\     
     Replace a line & \texttt{REPL} (replace line) \\  
     Add a typo to a word & \texttt{REPW} (replace word) \\  
     Randomly replace any character in a line & \texttt{REPL} (replace line) \\  
    \bottomrule
  \end{tabular}
\end{table}

We employ five types of corruptions in our pipeline and match each with a DSL command that can reverse it; see Table~\ref{tab:corruptions}. Some corruptions are potentially reversible by multiple DSL commands or a sequence of commands. For example, a corruption that adds a typo to a line of code may be inverted by replacing the word that contains the typo or by replacing the entire line with a corrected line. It may also be inverted by deleting the offending line and separately adding a corrected line, using two turns of interaction. Here, we opt to match every corruption type with a single local DSL command for simplicity.

To generate the synthetic dataset, we take tasks and ground-truth code snippets from the Mostly Basic Python Problems (MBPP) dataset~\citep{austin2021mbpp}. For each task in the MBPP dataset, we first sample the total number of corruptions to be applied between 1 and 5 (inclusive) and then sample that many corruption types uniformly with replacement from those in Table~\ref{tab:corruptions}. To obtain diverse trajectories, we repeat this procedure 100 times for each task in the train split of MBPP, and 10 times each in the validation split. This leads to a dataset of 35,223 demonstrations for training (19.7M tokens) and 889 for evaluation after de-duplication. We take 180 of the evaluation demonstrations for validation data, and use the remaining 709 as a test set. We fine-tune our models to imitate these demonstrations with supervised fine-tuning using LoRA, see Section~\ref{sec:training} for details.

\subsection{Reinforcement Learning with Verifiable Rewards on the code repair benchmark}
After training the model to use the text editor, we employ RLVR on the final code repair task, using a standard on-policy RL procedure. The code repair task consists of repairing incorrect solutions to problems from the MBPP training set. However, rather than being generated as a sequence of simple corruptions, these incorrect solutions are directly generated by a LM. This results in a much harder task, where initial `solutions' are often much further from correct than those in the CoE demonstrations. We outline how this code repair benchmark is constructed in Section~\ref{subsec:mbppr}.

The model is prompted to use the CoE procedure to repair these initial solutions. As in the previous SFT stage, the prompt describes the task and the usage of the DSL commands. This prompt is followed by the initial state, which contains the broken code snippet (equipped with line markers) and its corresponding execution feedback, as outlined in Section~\ref{subsec:scr}. At every interaction turn, the model is prompted with the CoE so far (i.e., all previous states and actions) and outputs an action in the form of an editing command. This action is executed in the scratchpad and rewarded using an engineered reward function. We store the resulting experience in a buffer and train with LoRA on an adapted version of the Group Relative Policy Optimization (GRPO) objective \citep{Shao2024-grpo}. Our method adapts the reward normalization in GRPO to use per-turn statistics, rather than the full trajectory statistics for all turns. This is similar to employing a variance-reducing \emph{return baseline} estimated from parallel rollouts of the RL policy, a strategy that has been successfully applied in the RL literature~\citep{kool2019buy, Kool2020Estimating, bakker2020single}. We refer to Appendix~\ref{app:rlft} for more details.

\paragraph{Reward design:} Our reward function is the sum of a task reward term and format reward term. We ran experiments using three different task rewards, described in Appendix~\ref{app:rlft}. In the end, the best-performing reward function simply rewards the model with a value of $1.0$ if and only if the task is solved in the current turn and was not solved in the previous turn. The format reward term applies a penalty of $-0.5$ when the model outputs an action that does not use the correct DSL syntax. Additionally, the model is penalized $-0.5$ when it does not output an \texttt{\small EXIT} action after the code snippet has been repaired (passes all unit tests).

\section{Experiments \label{sec:exp}}

\subsection{The code repair benchmark \label{subsec:mbppr}}
Our target code repair benchmark again consists of problems based on the MBPP dataset. We largely follow the procedure outlined in~\citet{Ni2024-mbppr}\footnote{It was our intention to directly use the MBPP-R benchmark contributed in this reference. However, we could not find a Github or dataset release for it, nor could we obtain it from the authors. We therefore decided to use their method to generate a new benchmark.}. For each MBPP problem, we 3-shot prompt \texttt{\small Llama-3.1-8B} (base model) to generate 100 solutions. Of these, we keep up to 20 that \emph{fail to solve} the task, while taking care that there are no repeats of the same proposed `solution'. Each of these, once associated with the original task description and ground-truth code snippet, defines a code repair task. We generate both a training and an evaluation split; in order to increase the size of our training split, and again following~\citet{Ni2024-mbppr}, we re-split the MBPP dataset and pool together its training split and half of its test split to make our training split; for our evaluation split, we use the tasks from the evaluation split of MBPP. In the end, we generate 9760 code repair tasks for the train split and 1497 for the evaluation split. The tasks in the train split are used for training during the RLVR stage. We take 180 evaluation split problems as a validation dataset for model selection, and the remaining 1317 problems are used as the test split for final evaluation. 

We refer to Appendix~\ref{app:mbppr} for more details on this dataset. Therein we also report two metrics: the average edit distance of the snippets of code to be repaired from their respective ground truth, and the 3-shot repair performance of all base (before training) models we consider on both the repair benchmark and synthetic corruptions we outlined in Section~\ref{sec:synth}.
Both metrics show that the tasks in our repair benchmark are significantly more difficult than those in the CoE demonstrations dataset, making it an appropriate test for our pipeline. Additionally, the ability to repair incorrectly LLM-generated initial code snippets is increasingly relevant, as LLMs are increasingly employed as coding agents, see e.g.~\citet{Gehring2024-rlef}.

\subsection{The natural language reasoning baseline}
Our CoEs are intended as a constrained agentic counterpart of CoTs, meant to enable SLMs to solve code repair tasks step by step without the verbosity of CoTs. We therefore compare our own training pipeline with one that is intended to enable SLMs to employ \emph{CoTs} on these same tasks. 
In order to do this, we separately SFT the pre-trained models that we use in our pipeline on the s1K dataset~\citep{Muennighoff2025-ks}\footnote{\url{https://huggingface.co/datasets/simplescaling/s1K}}, which consists of reasoning problems paired with high-quality CoTs.
In order to facilitate transfer from these problems to code repair, we design a prompt template that we employ both during fine-tuning on s1K and during evaluation. We refer to Appendix~\ref{app:s1} for details on s1K fine-tuning and the prompt template. 

\subsection{Training and evaluation setups} \label{sec:training}
Training (both SFT and RLVR) is done using LoRA~\citep{Hu2021-lora} and the AdamW optimizer \citep{loshchilov2018decoupled} on a family of \texttt{\small Llama} models of three sizes (1B, 3B, 8B). We employ rank 16 adapters on all linear layers except the encoding and decoding layers of \texttt{\small Llama-3.2-1B}, \texttt{\small Llama-3.2-3B}, and \texttt{\small Llama-3.1-8B-Instruct}. In order to facilitate training, we 4bit quantize the 8B model. During SFT on CoE demonstrations, we train on a standard next-token prediction objective on the full trace. Thus, the model is trained to predict not just the actions, but also the states, to directly learn about the DSL logic. During RLVR, we instead train only on the model-generated tokens (actions), using GRPO with a group size of 4, where every group consists of trajectories starting from the same initial prompt. When sampling experience, we use a temperature of $0.7$. We perform one epoch of gradient updates on that experience before discarding it, leading to functionally on-policy updates. We batch transitions such that training fits on GPU memory, and employ gradient accumulation to artificially increase batch size during both SFT and RLVR. All our experiments were performed on single GPU, with the 1B model fully trainable using only 16GB of GPU memory; see Appendix~\ref{app:training} for more details.

We report pass@1 and pass@4 rates for our CoE approach, the s1K baseline, and direct answer (i.e., no ItC) on the base models. We use greedy decoding to compute pass@1, and sample at a temperature of 0.2 to compute pass@4. During evaluation, models trained to output CoEs are coupled with the text editor, whose state is appended to the model's context after each editing command. 
In this way, the model is not required to predict the next state and observes accurate execution feedback at each step. The model is required to solve the task (fix the code snippet and \texttt{EXIT}) within 10 turns. Conversely, models trained to output CoTs are not coupled with the text editor and must end their answer with a correct snippet of Python code between pre-set delimiters; in order to succeed, they must do so while generating at most 2048 tokens.

\paragraph{Model selection:} During SFT on CoE demonstrations (step 1 of our pipeline), we evaluate checkpoints every 250 training steps on 180 tasks from the CoE validation dataset. For every model size (1B, 3B, 8B), we select the checkpoint that solves the most of these tasks as a starting point for RLVR training. During RLVR training (step 2 of our pipeline), we evaluate our models on an evaluation dataset of 180 code repair tasks every 10 training iterations, saving every checkpoint that performs on-par with or better than all previous checkpoints. For runs that perform well on the evaluation dataset, we select at most two checkpoints to evaluate on the test set of 1317 tasks and report the score of the best performing checkpoint. For the natural reasoning baseline, we evaluate on the test set after 1 and 5 epochs of training, and report the best performing checkpoint of the two; note that the s1K paper~\citet{Muennighoff2025-ks} prescribes training for 5 epochs. Notably, the 1-epoch checkpoint performs best for the 8B model, while the 5-epoch checkpoint performs best for the 1B model. The difference is negligible for the 3B model.

\subsection{Results}
We report our main results, comparing our RLVR pipeline with natural language reasoning trained on the s1k dataset and direct answer, in Table~\ref{tab:main_results}.
\begin{table}[h!]
    \caption{Main results comparing our RLVR pipeline with natural reasoning trained on s1k, for three different model sizes. Our \texttt{\small Llama-3.1-8B-Instruct} model is 4bit quantized. Best performing method per model and metric is highlighted in bold.}
    \label{tab:main_results}
    \centering
    \begin{tabular}{ccccccc}
        \toprule
        & \multicolumn{2}{c}{\textbf{Ours (CoE)}} & \multicolumn{2}{c}{\textbf{SFT on s1k (CoT)}} & \multicolumn{2}{c}{\textbf{Direct answer (3-shot)}} \\
        \cmidrule(r){2-3} \cmidrule(l){4-5} \cmidrule(l){6-7}
        \textbf{Model} & pass@1 & pass@4 & pass@1 & pass@4 & pass@1 & pass@4 \\
        \midrule
        \texttt{\small Llama-3.2-1B} & \textbf{7.82}\% & \textbf{11.0}\% & 0.15\% & 0.53\% & 1.3\% & 3.1\% \\
        \texttt{\small Llama-3.2-3B} & \textbf{13.8}\% & \textbf{19.0}\% & 1.44\% & 5.24\% & 6.9\% & 12.0\% \\
        \texttt{\small Llama-3.1-8B-Instruct} & 21.7\% & 32.7\% & 23.3\% & \textbf{46.2}\% & \textbf{33.4}\% & 42.9\% \\
        \bottomrule
    \end{tabular}
\end{table}

We observe large improvements using our CoE pipeline for the 1B and 3B models over the direct-answer baseline, on both the pass@1 and pass@4 metrics. The performance gap is especially notable for the smallest (1B) model. This suggests that our CoE method indeed allows these small language models to improve their code repair capabilities by utilizing the turn-based structure and execution feedback of our environment; interaction with this environment provides a way for small models to use additional tokens to improve their code repair performance beyond what is possible with immediate answering. We provide an instructive example of code repair trace for the 1B model in Box~\ref{box:trace_example} of the Appendix, with a brief discussion in Appendix~\ref{app:mbppr}. Furthermore, fine-tuning on s1K fails to instill improved code repair capabilities in these models based on ``natural language thinking'', instead leading to reduced performance on both metrics for the 1B and 3B models. Inspecting the model output for these s1K-trained models shows that they regularly get stuck repeating common patterns: output degenerates into printing consecutive line numbers, repeating reasoning sentences such as `\texttt{\small The condition for a triangle to be scalene is that none of the three sides are equal.}' or paragraphs, or outputting simulated execution feedback. Additionally, the small models often fail to use thinking and solution delimiters correctly.

For the larger 8B model, these observations reverse. Performance on the 3-shot baseline shoots up, and CoE training here seems to hamper the model's repair performance, which suggests that the model cannot utilize information it has learned during pre-training as effectively in a turn-by-turn setting as it can in a direct-answer setting. The s1K-trained model also performs much better here, and manual inspection of model output often shows coherent reasoning about specific failures of the initial code snippet. In one instance, the model encounters an MBPP task to sum the factors of a given integer, together with an initial code snippet that correctly performs this task but does not pass the unit tests. It then reasons backwards from the unit tests to determine that the actual task is to find \emph{prime} factors, and handily solves the task. While the s1K baseline does not outperform the direct answer baseline on the pass@1 metric, it does do so on pass@4, which further suggests that s1K LoRA fine-tuning adds useful diversity to generated outputs, even for models of this size (8B).


\section{Conclusions and limitations \label{sec:concl}}
In this paper, we have proposed an alternative way of parameterizing ``thinking'' tokens as a tool-usage trace, rather than as natural language reasoning. We show that this change allows us to extend the RLVR paradigm to models of size up to 3B, adapted exclusively via LoRA~\citep{Hu2021-lora}. The resulting agents are capable of using inference-time tokens in the form of environment interactions to improve their performance on a code-repair task; we moreover show that the same results cannot be obtained by attempting to elicit ``text-based'' reasoning in the form of natural-language CoTs. For the 8B model tested, this result reverses, suggesting that natural language reasoning via supervised fine-tuning on s1K may be feasible at this size.

\paragraph{Limitations:} We have focused on a code-repair task of our own making rather than more established reasoning benchmarks, and while we considered a range of (small) model sizes, we have confined our efforts to models of the \texttt{Llama} family. Adapting our CoE approach to more established reasoning or function-calling benchmarks, or to fixing snippets of code sourced from coding benchmarks richer than MBPP such as CodeContests \citep{li2022competition}, is a worthwhile direction for future research. Additionally, systematically benchmarking our CoE pipeline across a broad set of LMs of various sizes would make valuable follow-up work.

\bibliography{refs}
\bibliographystyle{abbrvnat}


\include{appendix}

\end{document}

%% file: appendix.tex
\newpage
\appendix

\section{Impact statement} \label{app:impact}
Recent developments in large-scale machine learning have made a definitive impact on society. Although such developments promise much positive impact in the form of new helpful technologies -- such as AI assistants, improvements in medical care, or mitigations to climate change, to name a few -- powerful technologies are inherently dual-use, and the scientific community should take care to address these. Whether the risks are enabling bad actors to do AI-assisted damage, harmful societal consequences due to large-scale misinformation or job losses, power concentration into the hands of a few institutions or persons with sole access to the best AI systems, or human obsolescence due to game-theoretic forces in an AI-driven economy or takover scenario, we must engage with the potential negative consequences of our research field. Most of the risks mentioned here seem driven by the capabilities and proliferation of the most powerful `frontier' models. Thus, we do not believe our current work meaningfully increases the risks associated with AI technologies, with perhaps the exception of risks associated with capabilities proliferation. We hope that our insights into improving reasoning in \emph{small} language models may help democratize access to AI capabilities, and that this effect offsets the potential risk of proliferation to bad actors.

\section{Data}

\subsection{Details on CoE demonstrations} \label{app:coe}
We employ specific string delimiters required by the scratchpad and verifier when constructing the CoE demonstrations: states and actions are separated by the delimiter \texttt{\small ;\textbackslash n}, which we term the `end of output' suffix, or \texttt{eoos}. The code snippet and execution feedback (that make up the state) are separated by \texttt{\small ***}, see Boxes~\ref{box:state_example1} and~\ref{box:state_example2} for some examples. These delimiters serve to increase (human) readability of the demonstrations and allow for controlling model generation during the later reinforcement learning stages, by using them as stop strings.
\begin{fancypromptbox}[label=box:state_example2]{Box \thetcbcounter: Example of a state, complete with line markers and execution feedback.}{\obeylines
\begin{minted}{python}
L  1 
L  2 def split_Arr(Arr,n,k):
L  3     x = k
L  4     y = n-k
L  5     Arr = Arr[:y]
L  6     Arr = Arr[::-1]
L  7     Arr = Arr + Arr[:x]
L  8     return Arr
\end{minted}
***
Test number 1 was not successful!
The code of the failed test was:
\mint{python}{assert split_Arr([12,10,5,6,52,36],6,2) == [5,6,52,36,12,10]}
Test number 2 was not successful!
The code of the failed test was:
\mint{python}{assert split_Arr([1,2,3,4],4,1) == [2,3,4,1]}
Test number 3 was not successful!
The code of the failed test was:
\mint{python}{assert split_Arr([0,1,2,3,4,5,6,7],8,3) == [3,4,5,6,7,0,1,2];}}
\end{fancypromptbox}


\paragraph{Limitations:} Note that the demonstrations we generate are not guaranteed to show the shortest possible sequence of edits from a corrupted to corrected code snippet. The sequence may, for instance, contain a subsequence that adds and then deletes the same line, since the randomly generated sequence of corruptions may have first added that line and subsequently deleted it. This is not a major issue, since the primary purpose of these synthetic demonstrations is to prepare the models for training with RLVR. A final limitation is that the \texttt{\small REPW} command, which replaces every instance of a word in a line with a different word, does not strictly reverse the typo-corruption with which it is matched. For example, the line \texttt{\small for i in range(10):} may be typo-corrupted into \texttt{\small for in in range(10):}. The command \texttt{\small REPW \#L \pyprefix in \pyprefix i} then recovers the line \texttt{\small for i i range(10):}, which is not the original line and yields a syntax error. This happens sufficiently rarely in our automated pipeline that we may simply skip the sequences for which this happens.

\paragraph{Corruption details:} When adding or replacing a line (\texttt{\small ADDL} and \texttt{\small REPL}), we randomly sample a line of code from a different ground truth code snippet in the current dataset split (training or validation) to insert into the current snippet. Typos (recovered by the \texttt{\small REPW} command) are generated by first randomly (uniform) selecting a word -- defined as any unit surrounded by whitespace -- in a specified line. Then, that word is replaced by a string that is selected uniformly at random from the set of strings that have Levenshtein-Damerau distance of 1 to the selected word.
\begin{table}
  \caption{The corruption types used in automated generation of \emph{Chain-of-Edits} demonstrations. Every corruption type is matched with a DSL command that can reverse the corruption.}
  \label{tab:corruptions}
  \centering
  \begin{tabular}{ll}
    \toprule
    Corruption type & Reversing DSL command \\
    \midrule
     Delete a line & \texttt{ADDL} (add line) \\
     Add a line & \texttt{DELL} (delete line) \\     
     Replace a line & \texttt{REPL} (replace line) \\  
     Add a typo to a word & \texttt{REPW} (replace word) \\  
     Randomly replace any character in a line & \texttt{REPL} (replace line) \\  
    \bottomrule
  \end{tabular}
\end{table}

\subsection{The code repair benchmark\label{app:mbppr}}
\begin{fancypromptbox}[label=box:mbppr_eval]{Box \thetcbcounter: Example code repair evaluation prompt for the \texttt{check\_isosceles} task.}{\obeylines
You are an expert Python programmer whose goal is to fix all mistakes in a code snippet. You may interact with the code snippet only by applying the provided DSL commands. Valid DSL command templates are:
\verb|`### DELL <line_number>`| to delete the line at the specified line number.
\verb|`### ADDL <line_number> >>><line_content>`| to add a line at the specified line number with the specified content.
\verb|`### REPL <line_number> >>><line_content>`| to replace the line at the specified line number with the specified content.
\verb|`### REPW <line_number> >>><string_to_replace> >>><string_to_insert>`| to replace all specified strings in the line at the line number with the new string.
Here is your task: \emph{Write a function to print check if the triangle is scalene or not.}
Your code should pass these tests:
\begin{minted}{python}
assert check_isosceles(6,8,12)==True
assert check_isosceles(6,6,12)==False
assert check_isosceles(6,15,20)==True
\end{minted}
Below is an initial malfunctioning code snippet to fix:
\begin{minted}{python}
L  1 
L  2 def check_isosceles(a, b, c):
L  3     if a == b or b == c or c == a:
L  4         return True
L  5     else:
L  6         return False
\end{minted}
*** 
Test number 1 was not successful!
The code of the failed test was:
\mint{python}{assert check_isosceles(6,8,12)==True}
Test number 2 was not successful!
The code of the failed test was:
\mint{python}{assert check_isosceles(6,6,12)==False}
Test number 3 was not successful!
The code of the failed test was:
\mint{python}{assert check_isosceles(6,15,20)==True;}}
\end{fancypromptbox}

As outlined in the main text, we generated the dataset by sampling candidate solutions of MBPP tasks from Llama3.1-8B. For each task, 100 solutions were sampled, with a sampling temperature of 0.8 to ensure sufficient diversity. From this 100, we filter out perfect duplicates and then accept at most 20, in order to avoid biasing the dataset towards difficult coding tasks. Therefore, for each MBPP task, we generate 20 or fewer code repair tasks. Box~\ref{box:mbppr_eval} shows an example of a resulting code repair task, which is formatted the same way as the CoE demonstrations.
\begin{table}[h!]
    \caption{Comparison of edit distance with ground truth MBPP solution for our two datasets. The repair dataset has much larger average edit distance as well as much more variation in edit distance over tasks in the dataset.}
    \label{tab:editdist}
    \centering
    \begin{tabular}{ccc}
        \toprule
        \textbf{Dataset} & Edit distance (mean) & Edit distance (stddev) \\
        \midrule
        CoE demonstrations & 39.93 & 27.87 \\
        Repair dataset & 135.6 & 132.6 \\
        \bottomrule
    \end{tabular}
\end{table}

These code snippets usually fail their task's unit tests, rather than failing due to more obvious errors in syntax or variable naming. We report a breakdown of the type of failures they undergo.
\begin{itemize}
\item Unit test failures: 81.2\%
\item Syntax errors: 2.6\%
\item Name Errors: 5.6\%
\item Wrong number or type of arguments: 4.8\%
\item Other errors: 5.4\%
\end{itemize}
These code repair problems are more difficult to solve than the tasks that appear in our CoE dataset, whose failures usually consist of syntax errors, incorrect variable names, and wrong package imports. As additional evidence that these code repair problems are much harder to solve than those in the CoE demonstrations, we report two more quantitative metrics. First, in Table~\ref{tab:editdist} we report the edit distance of the initial incorrect solutions in the test split of both datasets with their associated ground-truth code snippet in the MBPP dataset. This solution is one of the solutions to the task; a larger edit distance implies a more corrupted initial solution, which implies that the task is more difficult. Second, in Table~\ref{tab:few_shot} we report 3-shot `direct answer' results for the pre-trained \texttt{\small Llama} models on the test split of both datasets. The repair dataset results are those of the main Table~\ref{tab:main_results}. 3-Shot performance for initial code snippets from the CoE demonstrations dataset is much higher, indicating that these are far easier to repair for the pre-trained models.
\begin{table}[h!]
    \caption{Comparison of 3-shot evaluation performance of the three pre-trained models on our two datasets. The models have higher 3-shot performance for repairing initial code snippets from the CoE demonstrations dataset, indicating that these are easier to solve.}
    \label{tab:few_shot}
    \centering
    \begin{tabular}{ccccc}
        \toprule
        & \multicolumn{2}{c}{\textbf{CoE demonstrations (3-shot)}} & \multicolumn{2}{c}{\textbf{Repair dataset (3-shot)}} \\
        \cmidrule(r){2-3} \cmidrule(l){4-5}
        \textbf{Model} & pass@1 (\%) & pass@4 (\%) & pass@1 (\%) & pass@4 (\%) \\
        \midrule
        \texttt{\small Llama-3.2-1B} & 20.7 & 26.2 & 0.15 & 0.53 \\
        \texttt{\small Llama-3.2-3B} & 47.3 & 53.7 & 1.44 & 5.24 \\
        \texttt{\small Llama-3.1-8B-Instruct} & 72.8 & 78.6 & 23.3 & 46.2 \\
        \bottomrule
    \end{tabular}
\end{table}

Finally, we provide an example successful code repair trace generated by the trained 1B model in Box~\ref{box:trace_example}. Note that, although the final code snippet passes all unit tests and thus is considered correct, it technically does not fully align with the MBPP task; it fails to replace commas or dots with a colon in any input string, but the unit tests fail to cover for this scenario.
\begin{fancypromptbox}[label=box:trace_example]{Box \thetcbcounter: Example (short) code repair trace generated by the trained 1B model.}{\obeylines
You are an expert Python programmer whose goal is to fix all mistakes in a code snippet. You may interact with the code snippet only by applying the provided DSL commands. Valid DSL command templates are:
\verb|`### DELL <line_number>`| to delete the line at the specified line number.
\verb|`### ADDL <line_number> >>><line_content>`| to add a line at the specified line number with the specified content.
\verb|`### REPL <line_number> >>><line_content>`| to replace the line at the specified line number with the specified content.
\verb|`### REPW <line_number> >>><string_to_replace> >>><string_to_insert>`| to replace all specified strings in the line at the line number with the new string.
Here is your task: Write a function to replace all occurrences of spaces, commas, or dots with a colon in the given string by using regex.
Your code should pass these tests:
\begin{minted}{python}
assert fill_spaces('Boult Curve Wireless Neckband') ==
 'Boult:Curve:Wireless:Neckband'
assert fill_spaces('Stereo Sound Sweatproof') ==
 'Stereo:Sound:Sweatproof'
assert fill_spaces('Probass Curve Audio') == 'Probass:Curve:Audio'
\end{minted}
Below is an initial malfunctioning code snippet to fix:
\begin{minted}{python}
L  1 
L  2 import re
L  3 def fill_spaces(s):
L  4     s = re.sub(r'([,\.])', r':', s)
L  5     return s
\end{minted}
***
Test number 1 was not successful!
The code of the failed test was:
\begin{minted}{python}
assert fill_spaces('Boult Curve Wireless Neckband') ==
 'Boult:Curve:Wireless:Neckband'
\end{minted}
Test number 2 was not successful!
The code of the failed test was:
\begin{minted}{python}
assert fill_spaces('Stereo Sound Sweatproof') ==
 'Stereo:Sound:Sweatproof'
\end{minted}
Test number 3 was not successful!
The code of the failed test was:
\mint{python}{assert fill_spaces('Probass Curve Audio') == 'Probass:Curve:Audio';}
\verb|### REPL 4 >>>    s = s.replace (" ", ":");|
\begin{minted}{python}
L  1 
L  2 import re
L  3 def fill_spaces(s):
L  4     s = s.replace (" ", ":")
L  5     return s
\end{minted}
***
;
\verb|### EXIT;|}
\end{fancypromptbox}

\section{Training setup} \label{app:training}

\subsection{General design choices} \label{app:training_gen}
In this section, we discuss some design choices that apply to all experiments. We use the HuggingFace Transformers library \citep{wolf2020transf} to load \texttt{\small Llama-3.2-1B}, \texttt{\small Llama-3.2-3B} and \texttt{\small Llama-3.1-8B-Instruct} \citep{grattafiori2024llama3herdmodels}. We 4bit quantize the 8B model, due to GPU memory constraints. All training is done using rank-stabilized LoRA \citep{kalajdzievski2023rslora} adapters of rank 16.

For the non-quantized models, we train separate adapters at each stage in our pipeline, always merging adapters of previous training stages before staring the next round of training (however, we keep these adapters saved separately, for flexibility and to save storage space). For the quantized model, naively merging adapters is known to lead to performance drops, as the 32-bit weights in the LoRA adapters are reduced to 4-bit weights without an adaptation stage. Instead, we opt not to merge the adapters of the quantized 8B model in our pipeline; as a result, we continuously train the same adapter in all stages of our pipeline.

We refer to Table~\ref{tab:general_hyp} for a detailed breakdown of model settings.

\begin{table}
  \caption{Hyperparameters used for multi-turn supervised fine-tuning.}
  \begin{center}
  \begin{tabular}{lc}
    \toprule 
    Model quantization \\ 
    \midrule
    \texttt{\small Llama-3.2-1B} & No \\ 
    \texttt{\small Llama-3.2-3B} & No \\ 
    \texttt{\small Llama-3.1-8B-Instruct} & Yes \\   

    \midrule 
    Quantization parameters \\
    \midrule 
    Library used & \texttt{\small BitsAndBytes} \\ 
    Quantization type & 4bit \\
    4bit quantization type & \texttt{\small nf4} \\ 
    4bit compute type & \texttt{\small float16} \\ 
    4bit storage type & \texttt{\small float16} \\ 
    Double quantization & Yes \\
    
    \midrule
    LoRA parameters \\
    \midrule
    Rank & 16 \\
    Alpha / Effective alpha (after rank stabilization) & 64 / 16 \\
    Dropout & 0.1 \\
    Target modules & All linear layers (except encoding and decoding) \\
    Adapter weights type & \texttt{\small float32} \\

    \midrule
    Tokenizers \\
    \midrule 
    Padding side & Left \\
    Truncation side & Left \\
    \bottomrule
  \end{tabular}
  \label{tab:general_hyp}
  \end{center}
\end{table}




\subsection{Supervised Fine-Tuning on CoE demonstrations} \label{app:mtsft}
We train using rank 16 LoRA with the standard next-token prediction objective on all tokens in the CoE demonstration. Training proceeded remarkably stably as a function of hyperparameters. We experimented with learning rates in $\{10^{-5}, 5\cdot10^{-5}, 10^{-4}\}$ and training batch sizes in $\{2, 4, 8, 16\}$, but found no clear performance differences on average. We refer to Table~\ref{tab:mt_hyp} for a detailed breakdown of the hyperparameters used in our final `production' runs.

\begin{table}
  \caption{Hyperparameters used for multi-turn supervised fine-tuning.}
  \begin{center}
  \begin{tabular}{lc}
    \toprule
    General \\
    \midrule
    Number of training prompts total & 35223 \\
    Number of evaluation prompts total & 889 \\
    Number of evaluation prompts used during training & 180 \\
    Number of evaluation prompts used during testing & 709 \\
    Maximum context length during training (tokens) & 2048 \\
    Maximum number of environment turns (training and evaluation) & 10 \\
    Batch size & 2 \\ 
    Gradient accumulation steps & 2 \\
    GPU memory required & 16GB \\

    \midrule 
    Optimizer: AdamW \\
    \midrule 
    Learning rate & $5\cdot 10^{-5}$ \\
    Betas & (0.9, 0.999) \\
    Weight decay & 0.0 \\
    \bottomrule
  \end{tabular}
  \label{tab:mt_hyp}
  \end{center}
\end{table}

\subsection{Reinforcement Learning with Verifiable Rewards} \label{app:rlft}
In this section we give more details on our reinforcement learning setup. 

\paragraph{GRPO objective:} We train using a variant of the GRPO objective \citep{Shao2024-grpo}, adapted for multi-turn scenarios. GRPO was first introduced for single-turn settings, where a model is tasked with answering a question directly, after some number of reasoning tokens. In this setting, there is no natural notion of a `turn' or a `timestep'. Thus, it is sensible to normalize the (outcome or process) reward by using the statistics of all rewards observed \emph{throughout whole trajectories} in the group and computing the return as the sum of these normalized rewards (\citet{Shao2024-grpo}, Section 4.1.2 and 4.1.3). In our multi-turn setting, we may instead perform the group normalization of the return \emph{per turn}. This is similar to using a return baseline estimated from parallel rollouts of the RL policy, which has been a successful strategy in the RL literature \citep{kool2019buy, bakker2020single}. This baseline uses local (per turn) information rather than global (whole trajectory) information, which may improve its variance-reducing properties \citep{Kool2020Estimating}. Thus, we opt to first compute the return $R_t$ for turn $t\in[1, T]$ of a trajectory as the sum-of-rewards from that turn onward (discount factor $\gamma=1$). Then, we normalize these returns using \emph{per-turn} statistics of the group. In early development, we found that this change led to slightly improved stability during RL training. 

In particular, we employ the following GRPO objective to train our policy $\pi_\theta$:

\begin{alignat}{2} \label{eq:grpo}
    \mathcal{J}&_{GRPO}(\pi_\theta) = \mathbb{E}_{\bm{q} \sim p_Q, \{ \bm{o_i} \}_{i=1}^G \sim \pi_{\theta_{old}}(.|\bm{q})} \nonumber \\
    & \sum_{i=1}^G \sum_{t=1}^{T} \left\{ \text{min} \left[ \frac{\pi_\theta(o_{i,t}|\bm{q},\bm{o}_{i,<t})}{\pi_{\theta_{old}}(o_{i,t}|\bm{q},\bm{o}_{i,<t})} \hat{A}_{i,t}, \text{clip} \left( \frac{\pi_\theta(o_{i,t}|\bm{q},\bm{o}_{i,<t})}{\pi_{\theta_{old}}(o_{i,t}|\bm{q},\bm{o}_{i,<t})}, 1 - \epsilon, 1 + \epsilon \right) \hat{A}_{i,t}\right] \right. \\ 
    & \mspace{80mu} \left. - \; \beta \mathbb{D}_{KL} \left[ \pi_\theta || \pi_\text{ref} \right] \vphantom{\frac{1}{1}} \right\}, \nonumber
\end{alignat}

for groups of size $G$ on query $\bm{q} \sim p_Q$, where the advantage is given by

\begin{equation}
    \hat{A}_{i,t} = \frac{R(o_{i,t}|\bm{q}, \bm{o}_{i,<t}) - \hat{\mu}_t(\bm{q})}{\hat{\sigma}_t(\bm{q})},
\end{equation}

with $R(o_{i,t}|\bm{q}, \bm{o}_{i,<t})$ the return for observation $o_{i,t}$ given query $\bm{q}$ and previous observation $\bm{o}_{i,<t}$, and

\begin{equation}
    \hat{\mu}_t(\bm{q}) =  \frac{1}{G_t} \sum_{i \in G_t} R(o_{i,t}|\bm{q}, \bm{o}_{i,<t})\;, \;\;\;\;\; \hat{\sigma}_t(\bm{q}) = \frac{1}{G_t-1} \sum_{i \in G_t} \left[ R(o_{i,t}|\bm{q}, \bm{o}_{i,<t}) - \hat{\mu}_t(\bm{q}) \right]^2,
\end{equation}

where $G_t$ contains the indices of only those trajectories in the group that have not terminated at turn $t$. In this way, the normalization only uses return values observed for the group at the current turn. If $G_t$ contains only one trajectory (because all others have terminated), we omit the normalization step for that turn and simply use $\hat{A}_{i,t} = R(o_{i,t}|\bm{q}, \bm{o}_{i,<t})$. Note that Equation~\eqref{eq:grpo} employs one of the changes suggested in the \texttt{\small Dr.GRPO} algorithm of~\citet{liu2025drgrpo}: it does not divide the GRPO loss by the response token length $|\bm{o}_i|$. 

Each iteration, we take 4 queries from the training set and sample 4 (the GRPO group size) trajectories per query using $\pi_{\theta_{old}}$, scratchpad, and verifier. We store these trajectories in a buffer together with the log-probabilities of the associated actions under both the current policy $\pi_{\theta_{old}}$ and the reference model $\pi_\text{ref}$. We then perform \emph{one} epoch of mini-batched updates over this experience. Each mini-batch consists of effectively 6 turns of experience (using gradient accumulation to fit this batch in GPU memory). For each batch, we perform a single averaged update with the AdamW optimizer \citep{loshchilov2018decoupled} to maximize Equation~\ref{eq:grpo}.

\paragraph{Task reward function:} Here, we describe the task reward functions with which we experimented in more detail.
\begin{itemize}
    \item \texttt{\small OnlyWhenSolved}: the reward is $1.0$ if and only if the task is solved in the current turn and was not solved in the previous turn.
    \item \texttt{\small UnitTestFraction}: the reward is the change in the fraction of unit tests solved between the current turn and the last turn. This fraction is always between $0.0$ and $1.0$, so the reward is in $[-1.0, 1.0]$.
    \item \texttt{\small UnitTestFraction} + \texttt{\small EditDistanceBonus}: the reward is that of \texttt{\small UnitTestFraction} plus a bonus/penalty if the current code snippet looks more/less like the ground truth snippet than in the previous turn. This is measured as the change in fractional edit distance, using the \texttt{\small editdistance} Python package. This reward can fluctuate strongly for short code snippets, so we multiply this term by a small constant $0.1$ and clip the full reward in the range $[-10, 10]$.
\end{itemize}
The best performing RL runs all used the \texttt{\small OnlyWhenSolved} reward function. The combination of \texttt{\small UnitTestFraction} + \texttt{\small EditDistanceBonus} ranked second overall, while only using \texttt{\small UnitTestFraction} performed worst of the three.

\paragraph{Hyperparameters:} During development, we experimented with various task reward functions, format reward penalties in $[-2.0, 0.0]$, KL divergence weight values in $[0.0, 0.1]$, GRPO group sizes in $\{2, 4, 8\}$, learning rates in $[10^{-7}, 10^{-5}]$, experience sampling temperatures in $[0.2, 1.4]$, experience batch size in $\{4, 8, 16, 32\}$ and training batch sizes in $\{2, 4, 6, 8, 16\}$. We refer to Table~\ref{tab:rl_hyp} for a detailed breakdown of the final hyperparameters used for our `production' RL training runs.

\begin{table}
  \caption{Hyperparameters used for RL fine-tuning.}
  \begin{center}
  \begin{tabular}{lc}
    \toprule
    General \\
    \midrule
    Number of training prompts total & 4500 \\
    Number of evaluation prompts total & 180 \\
    Number of test prompts total & 1317 \\
    Number of training prompts seen during training & 4000 \\
    Maximum context length during training (tokens) & 2048 \\
    Maximum number of environment turns (training and evaluation) & 10 \\
    Maximum number of generated tokens per turn (training and evaluation) & 250 \\
    GPU memory used for training (1B model) & 16GB \\
    GPU memory used for training (3B and 8B models) & 64GB \\

    \midrule 
    Optimizer: AdamW \\
    \midrule 
    Learning rate & $2\cdot 10^{-6}$ \\
    Betas & (0.9, 0.999) \\
    Weight decay & 0.0 \\
    
    \midrule
    GRPO \\
    \midrule
    Trajectories sampled per iteration & 16 \\
    Sampling batch size & 8 \\
    Group size & 4 \\
    Training batch size (transitions) (1B model) & 2 \\
    Training batch size (transitions) (3B and 8B models) & 1 \\
    Gradient accumulation steps (1B model) & 3 \\
    Gradient accumulation steps (3B and 8B models) & 6 \\
    Discount factor $\gamma$ & 1.0 \\
    Clip value $\epsilon$ & 0.2 \\
    KL divergence weight $\beta$ & 0.01 \\
    Updates on each experience sample & 1 \\
    Experience sampling temperature & 0.7 \\ 
    Temperature for computing GRPO log probabilities & 1.0 \\
    Group normalization of returns & Per turn \\
    Subtract group mean return & Yes \\
    Divide returns by group standard deviation (opposed to \citep{liu2025drgrpo}) & Yes \\
    Average or sum log probabilities within a turn (following \citep{liu2025drgrpo}) & Sum \\ 

    \midrule
    Reward \\
    \midrule
    Task reward & \texttt{\small OnlyWhenSolved} \\
    Format reward penalty & -0.5 \\
    \bottomrule
  \end{tabular}
  \label{tab:rl_hyp}
  \end{center}
\end{table}


\section{Finetuning on s1K~\label{app:s1}}

\subsection{Prompt template}
When we finetune the model on the s1k dataset, we re-map the dataset using the prompt template in box~\ref{box:s1k_train_temp}, where \textcolor{blue}{\texttt{<question>}}, \textcolor{blue}{\texttt{<thinking  trajectories>}} and \textcolor{blue}{\texttt{<attempt>}} are the respective fields in the s1k dataset\footnote{\url{https://huggingface.co/datasets/simplescaling/s1K}}.
\begin{fancypromptbox}[label=box:s1k_train_temp]{Box \thetcbcounter: The prompt template we use during SFT on the s1K dataset.}{\obeylines
You are an expert and conscientious reasoner whose goal is to provide detailed answers to questions. You will now be provided with one such question; reason step-by-step about it, and format your reasoning as:
[BEGIN THINKING]
<reasoning in natural language>
[END THINKING]
When you are done thinking, output the answer to the question, formatted as: 
[BEGIN SOLUTION]
<answer in natural language>
[END SOLUTION]
The question to be answered is:
\textcolor{blue}{\texttt{<question>}}
[BEGIN THINKING]
\textcolor{blue}{\texttt{<thinking  trajectories>}}
[END THINKING]
[BEGIN SOLUTION]
\textcolor{blue}{\texttt{<attempt>}}
[END SOLUTION]}
\end{fancypromptbox}

During evaluation on code repair we use the same template, to facilitate transfer between the tasks in s1K and our code repair benchmark. Using the repair task shown in box~\ref{box:mbppr_eval} as an example, we report in box~\ref{box:s1k_eval} how a model fine-tuned on s1K would be prompted to solve it.
\begin{fancypromptbox}[label=box:s1k_eval]{Box \thetcbcounter: Example code repair evaluation prompt for the \texttt{check\_isosceles} task.}{\obeylines
You are an expert Python programmer whose goal is to fix all mistakes in a code snippet. What will follow is an outline of what the code snippet is supposed to do, given as a natural-language description followed by a list of 3 unit tests it is supposed to pass. Then you will be given the broken code snippet, along with the Python stack trace it generates when the unit tests are run. 
Here is your task: \emph{Write a function to print check if the triangle is scalene or not.}
Your code should pass these tests:
\begin{minted}{python}
assert check_isosceles(6,8,12)==True
assert check_isosceles(6,6,12)==False
assert check_isosceles(6,15,20)==True
\end{minted}
You will be provided with a malfunctioning snippet of code in python, followed by the resulting python stack trace. Reason step-by-step about it, the unit tests, the task, and the stack trace. Format your reasoning as:
[BEGIN THINKING]
<reasoning in natural language>
[END THINKING]
When you are done thinking, output the repaired code snippet, formatted as: 
[BEGIN SOLUTION]
<executable python code>
[END SOLUTION]
The code snippet to be fixed is:
\begin{minted}{python}
L  1 
L  2 def check_isosceles(a, b, c):
L  3     if a == b or b == c or c == a:
L  4         return True
L  5     else:
L  6         return False
\end{minted}
*** 
Test number 1 was not successful!
The code of the failed test was:
\mint{python}{assert check_isosceles(6,8,12)==True}
Test number 2 was not successful!
The code of the failed test was:
\mint{python}{assert check_isosceles(6,6,12)==False}
Test number 3 was not successful!
The code of the failed test was:
\mint{python}{assert check_isosceles(6,15,20)==True;}
[BEGIN THINKING]}
\end{fancypromptbox}

\subsection{Training and evaluation setup}
We fine-tune our models on s1K using a standard next-token prediction objective on training sequences formatted as demonstrated in box~\ref{box:s1k_train_temp}. Due to the length of some of the sequences (the longest is longer than 9000 tokens), we fix a batch size of one and use 4 steps of gradient accumulation to simulate a batch size of 4. We fine-tune for 5 epochs, following~\citet{Muennighoff2025-ks}. We report in table~\ref{tab:s1k_hyp} all the hyperparameters used. 
\begin{table}
  \caption{Hyperparameters used for supervised fine-tuning on s1K.}
  \begin{center}
  \begin{tabular}{lc}
    \toprule
    General \\
    \midrule
    Number of training prompts total & 1000 \\
    Number of evaluation prompts total & 0 \\
    Batch size & 1 \\ 
    Gradient accumulation steps & 4 \\
    GPU memory required & 16GB \\
    \midrule 
    Optimizer: AdamW \\
    \midrule 
    Learning rate & $5\cdot 10^{-5}$ \\
    Betas & (0.9, 0.999) \\
    Weight decay & 0.0 \\
    \bottomrule
  \end{tabular}
  \label{tab:s1k_hyp}
  \end{center}
\end{table}
When evaluating the finetuned models, we prompt them as shown in box~\ref{box:s1k_eval}, and simply grab and verify (i.e. run agains the unit tests) whatever the model generated between [BEGIN SOLUTION] and [END SOLUTION] delimiters. 

\section{3-shot evaluation of base models on code repair}
Our 3-shot prompt for evaluation of base (i.e. not yet fine-tuned) models on code repair is too long to fit in a single page. Because of this, we report in box~\ref{box:1shot_eval} a 1-shot version of it, again using the code repair task in box~\ref{box:mbppr_eval} as an example. When we evaluate base models, we propt them with the 3-shot prompt and then simply grab and verify (i.e. run agains the unit tests) whatever the model generated between [BEGIN SOLUTION] and [END SOLUTION] delimiters.
\begin{fancypromptbox}[label=box:1shot_eval, fontupper=\tiny]{Box \thetcbcounter: Example 1-shot code repair prompt for the \texttt{check\_isosceles} task.}{\obeylines
You are an expert Python programmer, and here is your task: Write a function to find the minimum cost path to reach (m, n) from (0, 0) for the given cost matrix cost[][] and a position (m, n) in cost[][].
Your code should pass these tests:
\begin{minted}{python}
assert min_cost([[1, 2, 3], [4, 8, 2], [1, 5, 3]], 2, 2) == 8
assert min_cost([[2, 3, 4], [5, 9, 3], [2, 6, 4]], 2, 2) == 12
assert min_cost([[3, 4, 5], [6, 10, 4], [3, 7, 5]], 2, 2) == 16
\end{minted}
Below is an initial malfunctioning code snippet to fix:
\begin{minted}{python}
L  1 def min_cost(cost, m, n): 
L  2   if m == 0 and n == 0: 
L  3     return 0
L  4   if m == 0: 
L  5     return cost[m][n]
L  6   if n == 0: 
L  7     return cost[m][n]
L  8   return min(cost[m][n],
                  min_cost(cost, m - 1, n),
                  min_cost(cost, m, n - 1))
\end{minted}
***
Test number 1 was not successful!
The code of the failed test was:
\mint{python}{assert min_cost([[1, 2, 3], [4, 8, 2], [1, 5, 3]], 2, 2) == 8}
Test number 2 was not successful!
The code of the failed test was:
\mint{python}{assert min_cost([[2, 3, 4], [5, 9, 3], [2, 6, 4]], 2, 2) == 12}
Test number 3 was not successful!
The code of the failed test was:
\mint{python}{assert min_cost([[3, 4, 5], [6, 10, 4], [3, 7, 5]], 2, 2) == 16}
The final correct Python function is:
[BEGIN SOLUTION]
\begin{minted}{python}
def min_cost(cost, m, n):
	tc = [[0 for x in range(C)] for x in range(R)]
	tc[0][0] = cost[0][0]
	for i in range(1, m+1):
		tc[i][0] = tc[i-1][0] + cost[i][0]
	for j in range(1, n+1):
		tc[0][j] = tc[0][j-1] + cost[0][j]
	for i in range(1, m+1):
		for j in range(1, n+1):
			tc[i][j] = min(tc[i-1][j-1],
                               tc[i-1][j],
                               tc[i][j-1]) + cost[i][j]
	return tc[m][n]
\end{minted}
[END SOLUTION]
You are an expert Python programmer, and here is your task: \emph{Write a function to print check if the triangle is scalene or not.}
Your code should pass these tests:
\begin{minted}{python}
assert check_isosceles(6,8,12)==True
assert check_isosceles(6,6,12)==False
assert check_isosceles(6,15,20)==True
\end{minted}
Below is an initial malfunctioning code snippet to fix:
\begin{minted}{python}
L  1 
L  2 def check_isosceles(a, b, c):
L  3     if a == b or b == c or c == a:
L  4         return True
L  5     else:
L  6         return False
\end{minted}
*** 
Test number 1 was not successful!
The code of the failed test was:
\mint{python}{assert check_isosceles(6,8,12)==True}
Test number 2 was not successful!
The code of the failed test was:
\mint{python}{assert check_isosceles(6,6,12)==False}
Test number 3 was not successful!
The code of the failed test was:
\mint{python}{assert check_isosceles(6,15,20)==True;}}
\end{fancypromptbox}